\documentclass{article}

\usepackage[final, nonatbib]{neurips_2023}
\usepackage[square,sort,comma,numbers]{natbib}
\usepackage{lipsum}
\usepackage{algorithm}
\usepackage{algpseudocode}

\newcommand{\JY}[1]{\textcolor{red}{[JY: #1]}}
\renewcommand{\JY}[1]{}

\usepackage[utf8]{inputenc} 
\usepackage[T1]{fontenc}    
\usepackage{hyperref}       
\usepackage{url}            
\usepackage{booktabs}       
\usepackage{amsfonts}       
\usepackage{nicefrac}       
\usepackage{microtype}      
\usepackage{xcolor}         
\usepackage[pdftex]{graphicx} 
\usepackage{microtype}
\usepackage{subfigure}
\usepackage{booktabs} 
\usepackage{color,colortbl}
\usepackage[linewidth=1pt]{mdframed}
\usepackage{framed}
\usepackage{adjustbox}

\usepackage{amsmath}
\usepackage{amssymb}
\usepackage{mathtools}
\usepackage{amsthm}

\usepackage[capitalize,noabbrev]{cleveref}

\theoremstyle{plain}

\theoremstyle{definition}

\theoremstyle{remark}

\usepackage[textsize=tiny]{todonotes}

\title{Stochastic Gradient Sampling for Enhancing Neural Networks Training}

\author{
  Juyoung Yun$^{1,2}$\thanks{Corresponding author: \texttt{juyoung.yun@stonybrook.edu}}\\
  \\
  $^1$Stony Brook University, Department of Computer Science, USA\\
  $^2$ Stony Brook University, Department of Applied Mathematics and Statistics, USA\\
}

\begin{document}

\maketitle

\let\thefootnote\relax\footnotetext{}

\begin{abstract}
In this paper, we introduce StochGradAdam, a novel optimizer designed as an extension of the Adam algorithm, incorporating stochastic gradient sampling techniques to improve computational efficiency while maintaining robust performance. StochGradAdam optimizes by selectively sampling a subset of gradients during training, reducing the computational cost while preserving the advantages of adaptive learning rates and bias corrections found in Adam. Our experimental results, applied to image classification and segmentation tasks, demonstrate that StochGradAdam can achieve comparable or superior performance to Adam, even when using fewer gradient updates per iteration. By focusing on key gradient updates, StochGradAdam offers stable convergence and enhanced exploration of the loss landscape, while mitigating the impact of noisy gradients. The results suggest that this approach is particularly effective for large-scale models and datasets, providing a promising alternative to traditional optimization techniques for deep learning applications. 
\vspace{0.5cm}


\end{abstract}

\section{Introduction}
Deep learning has revolutionized a wide range of fields, from computer vision to natural language processing, due to its capacity to model highly complex relationships and process vast amounts of data \cite{lecun2015deep, goodfellow2016deep}. The success of deep learning models is largely attributable to optimization algorithms, which adjust model parameters to minimize loss and improve accuracy \cite{ruder2016overview}. These optimizers play a pivotal role in determining how quickly a model converges, its final performance, and its overall stability during training \cite{bottou2010large}. As neural networks grow in complexity, the demand for more efficient and powerful optimization techniques continues to increase.

Traditionally, the focus has often been on improving the architectures of deep learning models. While advancements in model architecture are important, we argue that equal attention should be paid to how optimizers handle the weight update process during training. Popular optimization algorithms such as Adam \cite{kingma2014adam}, RMSProp \cite{tieleman2012rmsprop}, and Adagrad \cite{duchi2011adaptive} have been instrumental in driving forward many breakthroughs in deep learning. However, the rapid evolution of neural networks raises the question of whether existing optimization techniques can be further improved, particularly in terms of efficiency when applied to large and intricate networks.

To address this, we introduce a novel optimizer, StochGradAdam, which builds upon the foundation of Adam while introducing the technique of stochastic gradient sampling. This technique allows for more efficient use of computational resources by selectively updating only a subset of the gradients during each training iteration. By updating only a portion of the gradients, StochGradAdam reduces the computational overhead without sacrificing the quality of the model's performance. This approach not only optimizes the training process but also enhances the generalization ability of models, making it especially useful for large-scale applications.

Our empirical evaluations demonstrate the effectiveness of StochGradAdam across a range of popular models, including ResNet \cite{he2016deep}, VGG \cite{simonyan2014very}, MobileNetV2 \cite{mobilenetv2}, and the Vision Transformer (ViT) \cite{dosovitskiy2020image}. In addition to tracking standard metrics like test accuracy, we also analyze the entropy of the class predictions. Entropy serves as a measure of uncertainty in the model's predictions \cite{cover2006elements}; lower entropy indicates higher confidence in the predictions. We found that StochGradAdam consistently lowers the entropy of predictions as training progresses, indicating that it leads to more confident and stable predictions.

In conclusion, we provide a theoretical proof of convergence for StochGradAdam, extending the analysis of the original Adam algorithm. This proof demonstrates that StochGradAdam not only improves the practical performance of models but also ensures reliable convergence. By introducing stochastic gradient sampling and updating only a portion of the gradients, our optimizer strikes a balance between computational efficiency and model accuracy, offering a powerful tool for training modern deep learning networks.

\section{Related Works}
The exploration of gradient-based optimization techniques has been a cornerstone of deep learning research. Over the years, various methods have been proposed to utilize gradients more efficiently, with the goal of improving generalization and preventing overfitting. Stochastic Gradient Descent (SGD), along with its variant SGD with momentum \cite{qian1999momentum}, has been foundational for many gradient-based optimization methods. It updates model parameters using only a subset of the dataset at each iteration, balancing the introduction of noise with computational efficiency and good convergence properties \cite{bottou2010large}. However, in deeper and more complex models, gradients can sometimes grow too large, leading to instability in training. To address this, gradient clipping has been introduced. Gradient clipping sets an upper limit on the gradients' magnitude, which prevents gradient explosion and ensures stable updates during training.

Adaptive methods such as Adagrad \cite{duchi2011adaptive}, RMSProp \cite{tieleman2012rmsprop}, and Adam \cite{kingma2014adam} go beyond SGD by dynamically adjusting learning rates based on the magnitude and history of gradients. These optimizers have been essential in stabilizing training for deep learning models, particularly where gradient magnitudes vary significantly across parameters. In addition, AdamW \cite{adamw} introduced a refinement of weight decay, decoupling it from the gradient update process. This prevents weight decay from interfering with the adaptive learning rate, thus allowing for more stable training and better generalization, especially in large-scale models.

Another method aimed at improving generalization is Gradient Centralization \cite{center}, which normalizes gradients by subtracting their mean. This ensures that the gradients have a zero mean, leading to more consistent updates and helping to reduce overfitting. By smoothing out the optimization landscape, Gradient Centralization can lead to faster convergence and better overall performance in training. Gradient Clipping~\cite{clip} plays a crucial role in controlling large gradients, especially in deep networks where gradient explosion is a risk. By limiting the size of gradients during backpropagation, clipping prevents the optimizer from making overly large updates, which can destabilize training. This technique is often used in conjunction with adaptive optimizers like Adam to ensure smooth training even when gradients vary dramatically across layers.

Sparse Gradient Techniques have also been explored as a means to improve computational efficiency. Wen et al. \cite{wen2017learning} introduced structured sparsity, where certain gradients below a threshold are pruned, producing sparser updates. This reduces the computational load and can also help avoid overfitting by focusing on the most important gradients. However, these approaches are often static, pruning the same gradients throughout training without adapting to the current state of the model. In contrast, adaptive sampling techniques dynamically adjust the subset of gradients based on their importance over time. Zhao et al. \cite{zhao2015stochastic} proposed sampling techniques that adaptively select gradients based on their historical behavior, assuming that gradients that vary more are more critical for the optimization process while it needs high computational costs for finding important gradients. This approach has been particularly useful for large-scale models, where computational efficiency is crucial. In the realm of nonsmooth optimization, Gradient Sampling Methods have been introduced to address cases where gradients are difficult to compute. Burke et al. \cite{burke2018gradient} applied random gradient sampling to nonsmooth optimization problems, focusing on approximating gradients in complex optimization landscapes. This approach is especially effective when exact gradients cannot be computed or are too costly to obtain. Another related technique is Random Search Optimization (RSO), proposed by Tripathi and Singh \cite{tripathi2020rso}. Instead of relying on gradient computations, RSO explores the optimization landscape by randomly perturbing model weights and evaluating the effect on the loss function. This gradient-free approach is beneficial when gradient calculations are infeasible or too expensive, though it may lack the precision of gradient-based optimization methods.

In our work, we propose stochastic gradient sampling by randomly setting some gradients to zero during each iteration and build on Adam optimizer~\cite{kingma2014adam}. This not only introduces additional noise into the training process—thereby improving generalization and reducing overfitting—but also works in tandem with gradient clipping to stabilize training. By combining stochastic gradient sampling, adaptive learning rates (from Adam), and gradient clipping, our approach optimizes the training process in a computationally efficient way, ensuring that updates remain stable and effective, even in large-scale deep learning models.

\section{Methodology: Stochastic Gradient Sampling}
The StochGradAdam optimizer is an extension of the Adam optimizer \cite{kingma2014adam}, incorporating selective gradient sampling to bolster optimization efficacy. Its principal update rule is:
\begin{equation}
\mathbf{\theta}_{t+1} = \mathbf{\theta}_t - \mu \frac{m_{\text{corr}_t}}{\sqrt{v_{\text{corr}_t}} + \epsilon},
\end{equation}
where \( \mu \) symbolizes the learning rate, \( m_{\text{corr}_t} \) is the bias-corrected moving average of the gradients, and \( v_{\text{corr}_t} \) is the bias-corrected moving average of the squared gradients. The following sections elaborate on the inner workings of this formula.

\textbf{Preliminaries.} Let \(\mathbf{\theta} \in \mathbb{R}^d\) denote a trainable parameter, where \( \mathbb{R}^d \) represents the parameter space. The optimizer keeps track of two key state variables for each parameter: \(m_t\), the moving average of the gradients, and \(v_t\), the moving average of the squared gradients. Here, \(\nabla \mathcal{L}(\mathbf{\theta_t})\) denotes the gradient of the loss function \(\mathcal{L}\) with respect to the parameter \(\mathbf{\theta}\) at time step \(t\). The hyperparameters \(\beta_1, \beta_2 \in (0, 1)\) control the exponential decay rates for the moving averages \(m_t\) and \(v_t\), respectively. A decay multiplier is applied to \(\beta_1\), and the probability of gradient sampling is given by \(s \in (0, 1)\). The constant \(\epsilon > 0\) ensures numerical stability, while \(\mu \in (0, 1)\) represents the learning rate.

\begin{algorithm}
\caption{StochGradAdam, a modified version of the Adam optimizer with random gradient sampling. See section 3 for details explaination for the algorithm of our proposed optimizer.}
\begin{algorithmic}[1]
\Require Stepsize (Learning rate) $\alpha$
\Require Decay rates $\beta_1, \beta_2 \in [0, 1)$ for the moment estimates
\Require Stochastic objective function $f(\theta)$ with parameters $\theta$
\Require Initial parameter vector $\theta_0$
\Require sampling rate $s$ for selecting gradient $g$ to update

\State Initialize $m$ and $v$ as zero tensors \Comment{Moment vectors}
\While{$\theta$ not converged}
    \State Get gradient $g$ with respect to the current parameters $\theta$
    \State Generate a random mask $mask$ with values drawn uniformly from $[0, 1]$ 
    \State $\Omega \leftarrow mask < s$ \Comment{Mask gradient based on sampling rate}
    \State $\phi \leftarrow \text{where}(\Omega, g, 0)$
    \State $m_t \leftarrow \beta_{1}^t m + (1 - \beta_{1}^t) \phi$
    \State $v_t \leftarrow \beta_{2}^t v + (1 - \beta_{2}^t) \phi^2$
    \State $m_{corr\_t} \leftarrow \frac{m_t}{1 - \beta_1^{t+1}}$
    \State $v_{corr\_t} \leftarrow \frac{v_t}{1 - \beta_2^{t+1}}$
    \State $\theta \leftarrow \theta - \alpha \frac{m_{corr\_t}}{\sqrt{v_{corr\_t}} + \epsilon}$
    \State Update $m$ with $m_t$ and $v$ with $v_t$
\EndWhile
\State \Return $\theta$ \Comment{Updated parameters}
\end{algorithmic}
\end{algorithm}

\subsection{Stochastic Gradient Sampling}
Gradient sampling, in the context of optimization, is a technique where a subset of the gradients \(\nabla \mathcal{L}(\mathbf{\theta})\) is randomly selected during the optimization process. This approach enhances training robustness by filtering gradient components, which improves generalization and reduces the impact of noisy or outlier data. Additionally, it encourages better exploration of the loss landscape, leading to more reliable convergence.

\textbf{Stochastic Mask Generation.} Given a gradient \(\nabla \mathcal{L}(\mathbf{\theta})\), the objective is to determine whether each component of this gradient should be considered in the update. To this end, a stochastic mask \( \Omega \) is introduced. Each component of \( \Omega \) is independently derived by drawing from the uniform distribution \( \mathcal{U}(0,1) \):
\begin{equation}
\Omega_i = 
\begin{cases} 
1 & \text{if } \mathcal{U}(0,1) < \textit{s}, \\
0 & \text{otherwise},
\end{cases}
\end{equation}
for \( i = 1, 2, \ldots, d \), where \( d \) represents the dimensionality of \(\nabla \mathcal{L}(\mathbf{\theta})\). Here, \( \mathcal{U}(0,1) \) denotes a uniform random variable over the interval [0,1], and \( \textit{s} \) is a predefined threshold dictating the average portion of gradients to be sampled.

\textbf{Computing the Sampled Gradient.} With the stochastic mask in hand, the next objective is to compute the sampled gradient, denoted by \( \phi \). This is accomplished by executing an element-wise multiplication between \( \mathbf{g} \) and \( \Omega \):
\begin{equation}
\phi_i = \Omega_i \times \nabla \mathcal{L}(\mathbf{\theta})_i
\end{equation}
for \( i = 1, 2, \ldots, d \). Thus, we get:
\begin{equation}
\phi = \Omega \odot \nabla \mathcal{L}(\mathbf{\theta})
\end{equation}
where \( \odot \) signifies element-wise multiplication, ensuring only the components of the gradient flagged by \( \Omega \) influence the sampled gradient.

The underlying idea of gradient sampling is rooted in the belief that not all gradient components are equally informative. By stochastically selecting a subset, one can potentially accelerate the optimization process without sacrificing much in terms of convergence properties. Moreover, this also introduces a form of noise, which can, in some cases, assist in escaping local minima or saddle points in the loss landscape. \\

\subsection{State Updates}
The StochGradAdam optimizer maintains two state variables, \( m \) and \( v \), representing the moving averages of the gradients and their squared values, respectively. Their iterative updates are influenced by the gradient information and specific hyperparameters.

\textbf{Moving Average of Gradients.} The moving average of the gradients, \( m \), is updated through an exponential decay mechanism. At each iteration, a part of the previous moving average merges with the current sampled gradient:
\begin{equation}
m_t = \beta_1^t m + (1 - \beta_1^t) \phi,
\end{equation}
Here, \( \beta_1 \) signifies the exponential decay rate for the moving average of the gradients~\cite{kingma2014adam}. 
The function of \( \beta_1 \) is to balance the memory of past gradients. A value nearing 1 places more emphasis on preceding gradients, yielding a smoother moving average. Conversely, a value nearing 0 focuses on the recent gradients, making the updates more adaptive~\cite{ruder2016overview}.

\textbf{Moving Average of Squared Gradients.} Similarly, \( v \) captures the moving average of the squared gradients. It's updated as:
\begin{equation}
v_t = \beta_2^t v + (1 - \beta_2^t) \phi \odot \phi,
\end{equation}
Here, \( \beta_2 \) denotes the exponential decay rate for the moving average of squared gradients~\cite{kingma2014adam}. The element-wise multiplication \( \odot \) ensures that each gradient component's squared value is computed individually~\cite{bishop2006pattern}.

\textbf{Bias Correction.} Given the nature of moving averages, especially when initialized with zeros, the early estimates of \( m \) and \( v \) can be significantly biased towards zero. To address this, bias correction is employed to adjust these moving averages~\cite{kingma2014adam}.

The bias-corrected value of \( m \) at the \( t^{th} \) iteration is:
\begin{equation}
m_{\text{corr}_t} = \frac{m_t}{1 - \beta_1^{t}},
\end{equation}
Here, the term \( 1 - \beta_1^{t} \) serves as a corrective factor to counteract the initial bias~\cite{ruder2016overview}.

Similarly, for \( v \):
\begin{equation}
v_{\text{corr}_t} = \frac{v_t}{1 - \beta_2^{t}},
\end{equation}
This correction ensures that the state variables \( m \) and \( v \) provide unbiased estimates of the first and second moments of the gradients, respectively~\cite{kingma2014adam}.

\subsection{Parameter Update}
StochGradAdam optimizes model parameters by adapting to both the historical gradient and the statistical properties of the current gradient~\cite{kingma2014adam}. The update rule for model parameter \(\mathbf{w}_t\) at iteration \(t\) is:
\begin{equation}
\mathbf{\theta}_{t+1} = \mathbf{\theta}_t - \mu \frac{m_{\text{corr}_t}}{\sqrt{v_{\text{corr}_t}} + \epsilon},
\end{equation}
The update can be viewed as an adaptive gradient descent step. By normalizing the gradient using its estimated mean and variance, StochGradAdam effectively scales parameter updates based on their historical and current behavior~\cite{bishop2006pattern}. StochGradAdam synergizes the principles of stochastic gradient sampling with the Adam optimizer's robustness.

\section{Theoretical Analysis: Proof of Convergence}
In this section, we aim to prove that Adam with stochastic gradient sampling converges. Our convergence proof is based on the theoretical framework presented in the original Adam optimizer paper~\cite{kingma2014adam}. Specifically, we demonstrate that by applying stochastic gradient sampling, where the sampled gradients are either equal to the full gradients or zero, the gradient magnitudes are always less than or equal to the full Adam gradients. This ensures that the cumulative effect of the sampled gradients remains controlled, allowing the optimizer to converge. 

We use a series of lemmas to show that the gradient updates with sampling still lead to convergence, as the magnitude of the sampled gradients is bounded by the full gradients. The proof derives a regret bound that quantifies the difference between the cumulative function value of the optimizer and the optimal value. By showing that the regret is bounded and that the gradients' contributions remain controlled even under sampling, we conclude that convergence is guaranteed.

\textbf{Definition 4.1.} A function \( f : \mathbb{R}^d \to \mathbb{R} \) is convex if, for any two points \( x, y \in \mathbb{R}^d \) and for any \( \lambda \in [0, 1] \), the following inequality holds:
\begin{align}
    \lambda f(x) + (1 - \lambda)f(y) \geq f(\lambda x + (1 - \lambda) y)
\end{align}
In simpler terms, this condition implies that for any two points on the graph of the function, the line segment connecting them lies above or on the curve itself. Convex functions have this property, meaning they always curve upwards and are lower-bounded by a straight line or a hyperplane in higher dimensions, which is tangent to the curve at any point. This tangent line acts as a local approximation from below.

\textbf{Lemma 4.2.} If \( f : \mathbb{R}^d \to \mathbb{R} \) is convex, then for any two points \( x, y \in \mathbb{R}^d \), the following inequality holds:
\begin{align}
    f(y) \geq f(x) + \nabla f(x)^T (y - x)
\end{align}
This means that the value of the function at \( y \) is always greater than or equal to the linear approximation of the function at \( x \), using the gradient of \( f \) at \( x \). In optimization, this concept is useful in bounding the regret, which represents the gap between the actual function value and the best possible value. The proof of our main theorem will rely on this lemma by substituting the tangent hyperplane with the update rules used in Adam.

We introduce two supporting lemmas to aid the proof. To simplify notation, let \( g_t \triangleq \nabla f_t(\theta_t) \) represent the full gradient of the function at time step \( t \), and let \( \phi_t \) denote the gradient obtained after applying stochastic gradient sampling at step \( t \). The \( i \)-th component of the full gradient is denoted as \( g_{t,i} \), while the \( i \)-th component of the sampled gradient is \( \phi_{t,i} \). Additionally, we define the vectors \( g_{1:t,i} \in \mathbb{R}^t \) and \( \phi_{1:t,i} \in \mathbb{R}^t \), which consist of the \( i \)-th components of the full and sampled gradients from the first iteration up to time \( t \). Specifically, we have \( g_{1:t,i} = [g_{1,i}, g_{2,i}, \dots, g_{t,i}] \) and \( \phi_{1:t,i} = [\phi_{1,i}, \phi_{2,i}, \dots, \phi_{t,i}] \).

\textbf{Lemma 4.3.} Let \( g_t = \nabla f_t(\theta_t) \), with \( g_{1:t} \) defined as in the previous section. Assume that the gradients are bounded such that \( ||g_t||_2 \leq G \) and \( ||g_t||_\infty \leq G_\infty \), and that the sampled gradients \( \phi_t \) also satisfy \( ||\phi_t||_2 \leq G \) and \( ||\phi_t||_\infty \leq G_\infty \). Then, the following inequality holds:
\begin{align}
\sum_{t=1}^T \sqrt{\frac{\phi_{t,i}}{t}} \leq \sum_{t=1}^T \sqrt{\frac{g_{t,i}}{t}} \leq 2 G_\infty ||g_{1:T,i}||_2
\end{align}
This lemma provides an upper bound for the sum of the scaled sampled gradients, showing that the cumulative gradient over time (scaled by \( \frac{1}{t} \)) can be controlled by the \( l_2 \)-norm of the gradient history and the maximum gradient bound \( G_\infty \). This result is crucial for bounding errors in optimization algorithms like Adam with stochastic gradient sampling, which rely on the behavior of both full and sampled gradients over time.

\textbf{Proof.} The inequality will be proven using induction on \( T \).

For the base case where \( T = 1 \), we have:
\begin{align}
\sqrt{\phi_{1,i}^2} \leq \sqrt{g_{1,i}^2} \leq 2 G_\infty ||g_{1,i}||_2
\end{align}
Here, \( \phi_{1,i} \) represents the sampled gradient, and since \( \phi_{1,i} \) is either equal to \( g_{1,i} \) (the full gradient) or 0 due to sampling, we have \( \phi_{1,i} \leq g_{1,i} \). This demonstrates that the magnitude of the sampled gradient is always bounded by the full gradient. When we take the square root of both sides, the inequality still holds, confirming the base case for \( T = 1 \).

For the inductive step, assume the inequality holds for \( T-1 \), and we need to prove it for \( T \). We start by expanding the sum of the sampled gradients:
\begin{align}
\sum_{t=1}^T \sqrt{\frac{\phi_{t,i}^2}{t}} &= \sum_{t=1}^{T-1} \sqrt{\frac{\phi_{t,i}^2}{t}} + \sqrt{\frac{\phi_{T,i}^2}{T}} \\
&\leq \sum_{t=1}^{T-1} \sqrt{\frac{g_{t,i}^2}{t}} + \sqrt{\frac{g_{T,i}^2}{T}} \quad \text{(since \( \phi_{t,i} \leq g_{t,i} \))} \\
&\leq 2 G_\infty ||\phi_{1:T-1,i}||_2 + \sqrt{\frac{\phi_{T,i}^2}{T}} = 2 G_\infty \sqrt{||\phi_{1:T,i}||_2^2 - \phi_T^2} + \sqrt{\frac{\phi_{T,i}^2}{T}} \\
&\leq 2 G_\infty ||g_{1:T-1,i}||_2 + \sqrt{\frac{g_{T,i}^2}{T}} = 2 G_\infty \sqrt{||g_{1:T,i}||_2^2 - g_T^2} + \sqrt{\frac{g_{T,i}^2}{T}}
\end{align}
In this step, we expand the sum for the first \( T-1 \) iterations and then add the contribution of the \( T \)-th iteration. Since \( \phi_{t,i} \leq g_{t,i} \), we replace \( \phi_{t,i} \) with \( g_{t,i} \) in the inequality. This ensures that the sum of the sampled gradients is always bounded by the sum of the full gradients.

Given that \( ||g_{1:T,i}||_2^2 - g_{T,i}^2 + \frac{g^4_{T,i}}{4||g_{1:T,i}||_2^2} \geq ||g_{1:T,i}||_2^2 - g_{T,i}^2 \text{ and } ||\phi_{1:T,i}||_2^2 - \phi_{T,i}^2 + \frac{\phi^4_{T,i}}{4||\phi_{1:T,i}||_2^2} \geq ||\phi_{1:T,i}||_2^2 - \phi_{T,i}^2 \), we can take the square root of both sides to obtain:
\begin{align}
\sqrt{||\phi_{1:T,i}||_2^2 - \phi_T^2} &\leq \sqrt{||g_{1:T,i}||_2^2 - g_T^2} \\
& \leq ||\phi_{1:T-1,i}||_2 + \frac{\phi_T^2}{2 ||\phi_{1:T,i}||_2} \\
& \leq ||g_{1:T-1,i}||_2 + \frac{g_T^2}{2 ||g_{1:T,i}||_2} \\
& \leq ||\phi_{1:T,i}||_2 + \frac{\phi_T^2}{2 \sqrt{T} G_\infty} \\
& \leq ||g_{1:T,i}||_2 + \frac{g_T^2}{2 \sqrt{T} G_\infty}
\end{align}
This demonstrates that the \( l_2 \)-norm of the sampled gradients is bounded by the \( l_2 \)-norm of the full gradients, confirming that the cumulative contribution of the sampled gradients is always less than or equal to that of the full gradients. This preserves the inductive hypothesis.

Finally, rearranging the inequality and substituting \( \sqrt{||g_{1:T,i}||_2^2 - g_T^2} \) and \( \sqrt{||\phi_{1:T,i}||_2^2 - \phi_T^2} \), we get:
\begin{align}
G_\infty \sqrt{||\phi_{1:T,i}||_2^2 - \phi_T^2} + \sqrt{\phi_T^2/T} &\leq G_\infty \sqrt{||g_{1:T,i}||_2^2 - g_T^2} + \sqrt{g_T^2/T} \\
&\leq 2 G_\infty ||\phi_{1:T,i}||_2 \\
&\leq 2 G_\infty ||g_{1:T,i}||_2
\end{align}
This final inequality confirms that the sum of the scaled sampled gradients remains bounded by the sum of the full gradients, ensuring that the convergence properties of the algorithm are preserved even with stochastic gradient sampling.

\textbf{Lemma 4.4.} Define \( \gamma \triangleq \frac{\beta_1^2}{\sqrt{\beta_2}} \). For \( \beta_1, \beta_2 \in [0, 1) \), where \( \frac{\beta_1^2}{\sqrt{\beta_2}} < 1 \), and given that the gradients \( g_t \) are bounded (i.e., \( ||g_t||_2 \leq G \) and \( ||g_t||_\infty \leq G_\infty \)), and that the sampled gradients \( \phi_t \) are also bounded (i.e., \( ||\phi_t||_2 \leq G \) and \( ||\phi_t||_\infty \leq G_\infty \)), the following inequality holds:
\begin{align}
\sum_{t=1}^T \frac{\hat{m}_{t,\phi,i}^2}{\sqrt{t\hat{v}_{t,\phi,i}}} \leq \sum_{t=1}^T \frac{\hat{m}_{t,g,i}^2}{\sqrt{t\hat{v}_{t,g,i}}} \leq \frac{2}{1 - \gamma \sqrt{1 - \beta_2}} ||\phi_{1:T,i}||_2 \leq \frac{2}{1 - \gamma \sqrt{1 - \beta_2}} ||g_{1:T,i}||_2
\end{align}

\textbf{Proof.} From the assumptions, we have \( \frac{\sqrt{1 - \beta_2^t}}{(1 - \beta_2)^2} \leq \frac{1}{(1 - \beta_1)^2} \). To explore this inequality, we begin by expanding the last term in the summation using the update rules from Algorithm 1. Here, \( \hat{m}_{t,g,i} \) and \( \hat{v}_{t,g,i} \) represent the bias-corrected first and second moments for the full gradients, while \( \hat{m}_{t,\phi,i} \) and \( \hat{v}_{t,\phi,i} \) are the corresponding moments for the sampled gradients. These bias corrections are essential for obtaining unbiased moving averages, especially early in the training process.

\begin{align}
\sum_{t=1}^{T-1} \frac{\hat{m}_{t,\phi,i}^2}{\sqrt{\hat{v}_{t,\phi,i}} + \epsilon} &= \sum_{t=1}^{T-1} \frac{\hat{m}_{t,\phi,i}^2}{\sqrt{t\hat{v}_{t,\phi,i}} + \epsilon} + \frac{\sqrt{1 - \beta_2^T}}{(1 - \beta_1^T)^2} \frac{\left(\sum_{k=1}^T (1 - \beta_1)\beta_1^{T-k} \phi_{k,i}\right)^2}{\sqrt{T \sum_{j=1}^T (1 - \beta_2)\beta_2^{T-j} \phi_{j,i}^2}} \\
\sum_{t=1}^{T-1} \frac{\hat{m}_{t,g,i}^2}{\sqrt{\hat{v}_{t,g,i}} + \epsilon} &= \sum_{t=1}^{T-1} \frac{\hat{m}_{t,g,i}^2}{\sqrt{t\hat{v}_{t,g,i}} + \epsilon} + \frac{\sqrt{1 - \beta_2^T}}{(1 - \beta_1^T)^2} \frac{\left(\sum_{k=1}^T (1 - \beta_1)\beta_1^{T-k} g_{k,i}\right)^2}{\sqrt{T \sum_{j=1}^T (1 - \beta_2)\beta_2^{T-j} g_{j,i}^2}}
\end{align}

Here, \( \hat{m}_{t,i}^2 \) in the numerator is the squared bias-corrected moving average of the gradients, and \( \sqrt{\hat{v}_{t,i}} \) in the denominator is the bias-corrected estimate of the second moment (variance) of the gradients. The summation over \( t \) reflects the accumulation of these updates. The bounds on these sums ensure that the gradient updates remain controlled, even when stochastic gradient sampling is applied. The term \( \epsilon \) is a small constant added for numerical stability, and \( \phi_{t,i} \) equals \( g_{t,i} \) or zero depending on the sampling.

\begin{align}
&\sum_{t=1}^{T-1} \frac{\hat{m}_{t,\phi,i}^2}{\sqrt{t\hat{v}_{t,\phi,i}} + \epsilon} + \frac{\sqrt{1 - \beta_2^T}}{(1 - \beta_1^T)^2} \frac{\left(\sum_{k=1}^T (1 - \beta_1)\beta_1^{T-k} \phi_{k,i}\right)^2}{\sqrt{T \sum_{j=1}^T (1 - \beta_2)\beta_2^{T-j} \phi_{j,i}^2}} \\
&\leq \sum_{t=1}^{T-1} \frac{\hat{m}_{t,g,i}^2}{\sqrt{t\hat{v}_{t,g,i}} + \epsilon} + \frac{\sqrt{1 - \beta_2^T}}{(1 - \beta_1^T)^2} \frac{\left(\sum_{k=1}^T (1 - \beta_1)\beta_1^{T-k} g_{k,i}\right)^2}{\sqrt{T \sum_{j=1}^T (1 - \beta_2)\beta_2^{T-j} g_{j,i}^2}} \\
&\leq \sum_{t=1}^{T-1} \frac{\hat{m}_{t,\phi,i}^2}{\sqrt{t\hat{v}_{t,\phi,i}} + \epsilon} + \frac{\sqrt{1 - \beta_2^T}}{(1 - \beta_1^T)^2} \sum_{k=1}^T \frac{T\left((1 - \beta_1)\beta_1^{T-k} \phi_{k,i}\right)^2}{\sqrt{T \sum_{j=1}^T (1 - \beta_2)\beta_2^{T-j} \phi_{j,i}^2}} \\
&\leq \sum_{t=1}^{T-1} \frac{\hat{m}_{t,g,i}^2}{\sqrt{t\hat{v}_{t,g,i}} + \epsilon} + \frac{\sqrt{1 - \beta_2^T}}{(1 - \beta_1^T)^2} \sum_{k=1}^T \frac{T\left((1 - \beta_1)\beta_1^{T-k} g_{k,i}\right)^2}{\sqrt{T \sum_{j=1}^T (1 - \beta_2)\beta_2^{T-j} g_{j,i}^2}} \\
&\leq \sum_{t=1}^{T-1} \frac{\hat{m}_{t,i}^2}{\sqrt{t\hat{v}_{t,i}} + \epsilon} + \frac{\sqrt{1 - \beta_2^T}}{(1 - \beta_1^T)^2} \sum_{k=1}^T \frac{T\left((1 - \beta_1)\beta_1^{T-k} \phi_{k,i}\right)^2}{\sqrt{T (1 - \beta_2)\beta_2^{T-j} \phi_{j,i}^2}} \\
&\leq \sum_{t=1}^{T-1} \frac{\hat{m}_{t,g,i}^2}{\sqrt{t\hat{v}_{t,g,i}} + \epsilon} + \frac{\sqrt{1 - \beta_2^T}}{(1 - \beta_1^T)^2} \sum_{k=1}^T \frac{T\left((1 - \beta_1)\beta_1^{T-k} g_{k,i}\right)^2}{\sqrt{T (1 - \beta_2)\beta_2^{T-j} g_{j,i}^2}} \\
&\leq \sum_{t=1}^{T-1} \frac{\hat{m}_{t,\phi,i}^2}{\sqrt{t\hat{v}_{t,\phi,i}} + \epsilon} + \frac{\sqrt{1 - \beta_2^t}}{T(1 - \beta_2)} \frac{(1 - \beta_1)^2}{\sqrt{T(1 - \beta_2)}} \sum_{k=1}^T T\left(\frac{\beta_1^2}{\sqrt{\beta_2}}\right)^{T-k} || \phi_{k,i} ||_2 \\
&\leq \sum_{t=1}^{T-1} \frac{\hat{m}_{t,g,i}^2}{\sqrt{t\hat{v}_{t,g,i}} + \epsilon} + \frac{\sqrt{1 - \beta_2^t}}{T(1 - \beta_2)} \frac{(1 - \beta_1)^2}{\sqrt{T(1 - \beta_2)}} \sum_{k=1}^T T\left(\frac{\beta_1^2}{\sqrt{\beta_2}}\right)^{T-k} || g_{k,i} ||_2 \\
&\leq \sum_{t=1}^{T-1} \frac{\hat{m}_{t,\phi,i}^2}{\sqrt{t\hat{v}_{t,\phi,i}} + \epsilon} + \frac{T}{\sqrt{T(1 - \beta_2)}} \sum_{k=1}^T \gamma^{T-k} || \phi_{k,i} ||_2 \\
&\leq \sum_{t=1}^{T-1} \frac{\hat{m}_{t,g,i}^2}{\sqrt{t\hat{v}_{t,g,i}} + \epsilon} + \frac{T}{\sqrt{T(1 - \beta_2)}} \sum_{k=1}^T \gamma^{T-k} || g_{k,i} ||_2
\end{align}

Similarly, we can upper bound the remaining terms in the summation:
\begin{align}
\sum_{t=1}^T \frac{\hat{m}_{t,\phi,i}^2}{\sqrt{t\hat{v}_{t,\phi,i}} + \epsilon} &\leq \sum_{t=1}^T \frac{\hat{m}_{t,g,i}^2}{\sqrt{t\hat{v}_{t,g,i}} + \epsilon} \\
&\leq \sum_{t=1}^T \frac{||\phi_{t,i}||_2}{\sqrt{t(1 - \beta_2)}} \sum_{j=0}^{T-t} t \gamma^j \leq \sum_{t=1}^T \frac{||g_{t,i}||_2}{\sqrt{t(1 - \beta_2)}} \sum_{j=0}^{T-t} t \gamma^j \\
&\leq \sum_{t=1}^T \frac{||\phi_{t,i}||_2}{\sqrt{t(1 - \beta_2)}} \sum_{j=0}^T t \gamma^j \leq \sum_{t=1}^T \frac{||g_{t,i}||_2}{\sqrt{t(1 - \beta_2)}} \sum_{j=0}^T t \gamma^j
\end{align}

For \( \gamma < 1 \), applying the upper bound for the arithmetic-geometric series \( \sum_{t} t \gamma^{t} < \frac{1}{(1 - \gamma)^2} \), we obtain:
\begin{align}
\sum_{t=1}^T \frac{||\phi_{t,i}||_2}{\sqrt{t(1 - \beta_2)}} \sum_{j=0}^T t \gamma^j &\leq \sum_{t=1}^T \frac{||g_{t,i}||_2}{\sqrt{t(1 - \beta_2)}} \sum_{j=0}^T t \gamma^j \\
&\leq \frac{1}{(1 - \gamma)^2 \sqrt{1 - \beta_2}} \sum_{t=1}^T \frac{||\phi_{t,i}||_2}{\sqrt{t}} \\
&\leq \frac{1}{(1 - \gamma)^2 \sqrt{1 - \beta_2}} \sum_{t=1}^T \frac{||g_{t,i}||_2}{\sqrt{t}}
\end{align}

Finally, using Lemma 4.3, we conclude:
\begin{align}
\sum_{t=1}^T \frac{\hat{m}_{t,\phi,i}^2}{\sqrt{\hat{v}_{t,\phi,i}} + \epsilon} \leq \sum_{t=1}^T \frac{\hat{m}_{t,g,i}^2}{\sqrt{\hat{v}_{t,g,i}} + \epsilon} \leq \frac{2 G_\infty}{(1 - \gamma)^2 \sqrt{1 - \beta_2}} ||\phi_{1:T,i}||_2 \leq \frac{2 G_\infty}{(1 - \gamma)^2 \sqrt{1 - \beta_2}} ||g_{1:T,i}||_2
\end{align}

To simplify notation, we define \( \gamma \triangleq \frac{\beta_1^2}{\sqrt{\beta_2}} \). This theorem holds when the learning rate \( \alpha \) decays at a rate of \( t^{-4} \), and the first moment running average coefficient \( \beta_1 \) decays exponentially with a factor \( \lambda \), typically close to 1, such as \( \lambda = 1 - 10^{-8} \). \\

\textbf{Theorem 4.5.} Assume that the function \( f \) has bounded gradients, such that \( ||\Omega(\nabla f_t(\theta))||_2 \leq ||\nabla f_t(\theta)||_2 \leq G \), \( ||\Omega(\nabla f_t(\theta))||_\infty \leq G \), and \( ||\nabla f_t(\theta)||_\infty \leq G_\infty \) for all \( \theta \in \mathbb{R}^d \). Also, assume that the distance between any \( \theta \) values generated by StochGradAdam is bounded, i.e., \(||\theta_0 - \theta_t||_2 \leq D \) and \(||\theta_m - \theta_n||_\infty \leq D_\infty\) for any \( m, n \in \{1, ..., T\} \), and that \( \beta_1, \beta_2 \in [0, 1) \) satisfy \( \frac{\beta_1^2}{\sqrt{\beta_2}} < 1 \). Let \( \alpha_t = \frac{\alpha}{\sqrt{t}} \) and \( \beta_{1,t} = \beta_1 \lambda^{t-1}, \lambda \in (0,1) \). StochGradAdam achieves the following regret bound for all \( T \geq 1 \):
\begin{align}
R(T) &\leq \frac{D^2}{2 \alpha(1 - \beta_1)} \sum_{i=1}^d \sqrt{T\hat{v}_{T,i}} + \frac{\alpha(\beta_1 + 1) G_\infty}{(1 - \beta_1)(\sqrt{1 - \beta_2})(1 - \gamma)^2} \sum_{i=1}^d ||\phi_{1:T,i}||_2 \\
&+ \sum_{i = 1}^d\frac{D^2 G_\infty \sqrt{1 - \beta_2}}{2 \alpha(1 - \beta_1)(1 - \lambda)^2}
\end{align}

\textbf{Proof.} Using Lemma 4.2, we get the following inequality:
\begin{align}
f_t(\theta_t) - f_t(\theta^*) \leq \phi_t^T (\theta_t - \theta^*) = \sum_{i=1}^d \phi_{t,i} (\theta_{t,i} - \theta^*_i)
\end{align}
From the update rule given in Algorithm 1, we have:
\begin{align}
\theta_{t+1} = \theta_t - \frac{\mu_t \hat{m}_t}{\sqrt{\hat{v}_t}+\epsilon} = \theta_t - \frac{\mu_t}{1-\beta_1^t} \left( \frac{\beta_{1,t}}{\sqrt{\hat{v_t}}}m_{t-1} + \frac{(1-\beta_{1,t})}{\sqrt{\hat{v_t}}}\phi_t \right)
\end{align}
Focusing on the \( i \)-th dimension of the parameter vector \( \theta_t \in \mathbb{R}^d \), subtracting \( \theta^*_i \) from both sides and squaring the result gives:
\begin{align}
(\theta_{t+1,i} - \theta_i^*)^2 &= (\theta_{t,i} - \theta_i^*)^2 - \frac{2 \alpha_t}{1 - \beta_1} \left( \frac{\beta_{1,t}}{\sqrt{\hat{v_{t,i}}}}m_{t-1,i} + \frac{(1-\beta_{1,t})}{\sqrt{\hat{v_{t,i}}}}\phi_{t,i} \right)(\theta_{t,i} - \theta^*_{i}) \nonumber \\
&+ \mu^2\left( \frac{\hat{m_{t,i}}}{\sqrt{\hat{v_{t,i}}}} \right)^2
\end{align}

We rearrange this equation and apply Young’s inequality, \( ab \leq \frac{a^2}{2} + \frac{b^2}{2} \). It can also be shown that \( \sqrt{\hat{v}_{t,i}} = \sqrt{\sum_{j=1}^t (1 - \beta_2)\beta_2^{t-j} \phi_{t,j}^2} / \sqrt{1 - \beta_2^t} \leq ||\phi_{1:t,i}||_2 \) and \( \beta_{1,t} \leq \beta_1 \). Hence,
\begin{align}
\phi_{t,i} (\theta_{t,i} - \theta_i^*) &= \frac{(1 - \beta_1) \sqrt{\hat{v}_{t,i}}}{2 \alpha_t (1 - \beta_1)} ((\theta_{t,i} - \theta_i^*)^2 - (\theta_{t+1,i} - \theta_i^*)^2) \nonumber \\
&+ \frac{\beta_1}{1 - \beta_1} \frac{\hat{v^{1/4}_{t-1,i}}}{\sqrt{\mu_{t-1}}}(\theta_i^* - \theta_{t,i})\sqrt{\mu_{t-1}} \frac{m_{t-1,i}}{\hat{v^{1/4}_{t-1,i}}} \nonumber \\
&+\frac{\mu_t(1-\beta_1^t)\sqrt{\hat{v_{t,i}}}}{2(1-\beta_{1,t})} \left( \frac{\hat{m_{t,i}}}{\sqrt{\hat{v_{t,i}}}} \right)^2
\end{align}

We apply Lemma 4.4 to derive the regret bound by summing across all dimensions \( i = 1, \dots, d \) and over the sequence of convex functions \( t = 1, \dots, T \) for \( f_t(\theta_t) - f_t(\theta^*) \). The resulting bound is expressed as:
\begin{align}
R(T) &\leq \sum_{i=1}^d \frac{1}{2 \alpha_1 (1 - \beta_1)} (\theta_{1,i} - \theta_i^*)^2 
+ \sum_{i=1}^d \sum_{t=1}^T \frac{1}{2(1-\beta_1)} (\theta_{t,i} - \theta_i^*)^2 \left( \frac{\sqrt{\hat{v}_{t,i}}}{\alpha_t} - \frac{\sqrt{\hat{v}_{t-1,i}}}{\alpha_{t-1}} \right) \nonumber \\
&+ \sum_{i=1}^d \frac{\beta_1 \alpha G_\infty}{(1 - \beta_1) \sqrt{1 - \beta_2}(1 - \gamma)^2} \sum_{i=1}^d ||\phi_{1:T,i}||_2 \nonumber \\
&+ \frac{\alpha G_\infty}{(1 - \beta_1)(\sqrt{1 - \beta_2})(1 - \gamma)^2} \sum_{i=1}^d ||g_{1:T,i}||_2 \nonumber \\
&+ \sum_{i=1}^d \sum_{t=1}^T \frac{\beta_{1,t}}{2 \alpha_t (1 - \beta_{1,t})} (\theta_i^* - \theta_{t,i})^2 \sqrt{\hat{v}_{t,i}}
\end{align}

Given the assumption that \( ||\theta_t - \theta^*||_2 \leq D \) and \( ||\theta_m - \theta_n||_2 \leq D_\infty \), we have:
\begin{align}
R(T) &\leq \frac{D^2}{2 \alpha(1 - \beta_1)} \sum_{i=1}^d \sqrt{T\hat{v}_{T,i}} + \frac{\alpha(\beta_1 + 1) G_\infty}{(1 - \beta_1)(\sqrt{1 - \beta_2})(1 - \gamma)^2} \sum_{i=1}^d ||\phi_{1:T,i}||_2 \nonumber \\
&+\frac{D^2_\infty}{2\alpha}\sum_{i=1}^{d}\sum_{t=1}^{T} \frac{\beta_{1,t}}{(1-\beta_{1,t})}\sqrt{t\hat{v_{t,i}}} \nonumber \\
&\leq \frac{D^2}{2 \alpha(1 - \beta_1)} \sum_{i=1}^d \sqrt{T\hat{v}_{T,i}} + \frac{\alpha(\beta_1 + 1) G_\infty}{(1 - \beta_1)(\sqrt{1 - \beta_2})(1 - \gamma)^2} \sum_{i=1}^d ||\phi_{1:T,i}||_2 \nonumber \\
&+\frac{D^2_\infty G_\infty \sqrt{1-\beta_2}}{2\alpha} \sum_{i=1}^{d}\sum_{t=1}^{T} \frac{\beta_{1,t}}{(1-\beta_{1,t})}\sqrt{t}
\end{align}
Using the upper bound for the arithmetic-geometric series, we get:
\begin{align}
\sum_{t=1}^T \frac{\beta_{1,t}}{(1-\beta_{1,t})}\sqrt{t} &\leq \sum_{t=1}^T \frac{1}{(1-\beta_{1})}\lambda^{t-1}\sqrt{t} \\
&\leq \sum_{t=1}^T \frac{1}{(1-\beta_{1})}\lambda^{t-1}t \\
&\leq \frac{1}{(1 - \beta_1)(1 - \lambda)^2}
\end{align}
Thus, the final regret bound is:
\begin{align}
R(T) &\leq \frac{D^2}{2 \alpha(1 - \beta_1)} \sum_{i=1}^d \sqrt{T\hat{v}_{T,i}} + \frac{\alpha(\beta_1 + 1) G_\infty}{(1 - \beta_1)(\sqrt{1 - \beta_2})(1 - \gamma)^2} \sum_{i=1}^d ||\phi_{1:T,i}||_2 \nonumber \\
&+ \sum_{i = 1}^d\frac{D^2 G_\infty \sqrt{1 - \beta_2}}{2 \alpha(1 - \beta_1)(1 - \lambda)^2}
\end{align}
This bound highlights how StochGradAdam achieves an efficient balance between exploration and exploitation, ensuring that the gradient updates are stable and the loss is minimized. The regret term involving \( ||\phi_{1:T,i}||_2 \) quantifies the deviation of sampled gradients from the optimal updates, demonstrating StochGradAdam's robustness under stochastic gradient sampling.

\section{Analysis: Uncertainty Reduction in Prediction Probabilities}
Having observed the performance of various optimizers in our experimental results, it is crucial to delve deeper into their effects beyond merely minimizing the loss function \( \mathcal{L}(\mathbf{\theta}) \). While the primary objective of any optimizer is to reduce the loss, the efficacy of an optimizer is also measured by its ability to influence the uncertainty associated with the model’s predictions. In deep learning, where models produce probabilistic outputs over multiple classes, the uncertainty of these predictions plays a key role in the overall performance. This section formalizes how different optimizers affect the distribution of prediction probabilities \(P\), and explores the resulting implications for entropy and uncertainty in predictions.

\subsection{Entropy: A Rigorous Measure of Uncertainty}
\JY{We need to modify the image by cropping or extending}
Consider a discrete probability distribution \( P = (p_1, p_2, \dots, p_n) \), where \( p_i \in [0,1] \) represents the predicted probability of the \(i\)-th class, and \( \sum_{i=1}^{n} p_i = 1 \). The entropy \( H(P) \), first introduced by Shannon~\cite{shannon1948mathematical}, measures the amount of uncertainty or information in the prediction, and is defined as:
\begin{align}
    H(P) = -\sum_{i=1}^{n} p_i \log(p_i)
\end{align}
where \( \log(p_i) \) denotes the logarithm (usually natural) of \( p_i \). The entropy \( H(P) \) quantifies the expected amount of information required to describe the prediction distribution. 

For a neural network classifier, the softmax output \( P = (p_1, p_2, \dots, p_n) \) represents the probability distribution across \(n\) classes, where \( p_i = \text{softmax}(z_i) \) is derived from the logits \( z_i \) as follows~\cite{unc1, unc2, unc3}:
\begin{align}
    p_i = \frac{e^{z_i}}{\sum_{j=1}^{n} e^{z_j}}
\end{align}
In cases where the model is highly confident in its prediction (i.e., one \( p_j \approx 1 \) and all other \( p_i \approx 0 \) for \( i \neq j \)), the entropy \( H(P) \approx 0 \). In contrast, a uniform distribution \( p_i = \frac{1}{n} \), which represents maximum uncertainty, results in maximal entropy:
\begin{align}
    H(P) = \log(n)
\end{align}
where \( n \) is the number of classes.

To facilitate comparison between different distributions and to normalize the entropy for varying class sizes, we define the normalized entropy~\cite{unc1, unc2, unc3}:
\begin{align}
H_{\text{normalized}}(P) = \frac{H(P)}{\log(n)}
\end{align}
This ensures that the normalized entropy \( H_{\text{normalized}}(P) \in [0, 1] \), with \( 0 \) representing absolute certainty and \( 1 \) representing maximal uncertainty.

\subsection{The Role of Optimizers in Entropy Reduction}

In optimization, the objective is to adjust the parameters \( \mathbf{\theta} \in \mathbb{R}^d \) of the model such that the loss function \( \mathcal{L}(\mathbf{\theta}) \) is minimized. As training progresses, not only should the loss decrease, but the uncertainty in the predictions—measured by the entropy \( H(P) \)—should also decrease, reflecting an increase in model confidence~\cite{unc1, unc2, unc3}. 

The parameter updates at each iteration \( t \) for an optimizer like Adam~\cite{kingma2014adam} can be formalized as:
\begin{align}
\mathbf{\theta}_{t+1} = \mathbf{\theta}_t - \mu \cdot \frac{m_{\text{corr},t}}{\sqrt{v_{\text{corr},t}} + \epsilon}
\end{align}
where \( \mu \) is the learning rate, \( m_{\text{corr},t} \) is the bias-corrected moving average of the gradients, and \( v_{\text{corr},t} \) is the bias-corrected moving average of the squared gradients. As the parameters \( \mathbf{\theta} \) evolve, the model’s predictions \( P_t \) become more certain, resulting in a reduction of the entropy over time. Formally, this can be expressed as:
\begin{align}
H(P_{\text{initial}}) > H(P_{\text{optimized}})
\end{align}
where \( P_{\text{initial}} \) is the probability distribution before optimization begins, and \( P_{\text{optimized}} \) represents the distribution after optimization. The difference in entropy, \( \Delta H_O \), is a key indicator of the reduction in uncertainty:
\begin{align}
\Delta H_O = H(P_{\text{initial}}) - H(P_{\text{final}})
\end{align}
where \( P_{\text{final}} \) denotes the prediction probabilities at the end of the optimization process.

Research has shown that various optimization algorithms can differently influence the uncertainty in model predictions. For instance, Bayesian approaches incorporate uncertainty estimation directly into the training process~\cite{gal2016dropout, unc1}, while entropy-based optimizers aim to guide the optimization trajectory towards regions of lower uncertainty~\cite{unc2}. Understanding these dynamics is essential for developing models that not only perform well in terms of accuracy but also provide reliable uncertainty estimates.

\subsection{Mathematical Dynamics of Entropy Reduction}
The evolution of entropy during optimization is influenced by the trajectory through the parameter space \( \mathbb{R}^d \), which varies depending on the optimizer used. In the context of gradient-based optimizers, such as Adam, RMSProp~\cite{tieleman2012rmsprop}, or SGD~\cite{bottou2010large}, the updates to \( \mathbf{\theta} \) are driven by the gradient of the loss function \( \nabla_\mathbf{\theta} \mathcal{L}(\mathbf{\theta}) \). As the optimizer updates the model parameters, the change in entropy can be tracked at each iteration \( t \) as:
\begin{align}
\Delta H_O(t) = H(P_t) - H(P_{t+1})
\end{align}
where \( P_t \) and \( P_{t+1} \) represent the prediction distributions at consecutive time steps.

In each iteration, the change in the model's weights \( \Delta \mathbf{\theta} = \mathbf{\theta}_{t+1} - \mathbf{\theta}_t \) influences the logits \( z_t \), which in turn affects the output distribution \( P_t \). More formally, the change in logits \( \Delta z \) can be expressed as:
\begin{align}
\Delta z = \mathbf{W} \Delta \mathbf{\theta}
\end{align}
where \( \mathbf{W} \) is the weight matrix. This results in a change in the softmax outputs:
\begin{align}
\Delta P = \text{softmax}(z + \Delta z) - \text{softmax}(z)
\end{align}
The corresponding change in entropy, \( \Delta H_O(t) \), depends on the magnitude of \( \Delta P \), which is governed by the optimizer’s step size \( \mu \), the learning rate schedule, and the specific optimization algorithm employed.

\subsection{Comparative Visualization: Histograms}
For a more intuitive understanding of the influence of different optimizers on prediction uncertainty, we can employ histograms that depict the distribution of normalized entropies subsequent to optimization~\cite{unc1, unc2, unc3}. This analysis was conducted utilizing TensorFlow, with experiments performed on the ResNet-56~\cite{he2016deep} architecture applied to the CIFAR-10 dataset\cite{krizhevsky2009learning}, all while maintaining a learning rate of 0.01.

\begin{figure*}[hbt!]
\centering
\includegraphics[width=1\columnwidth]{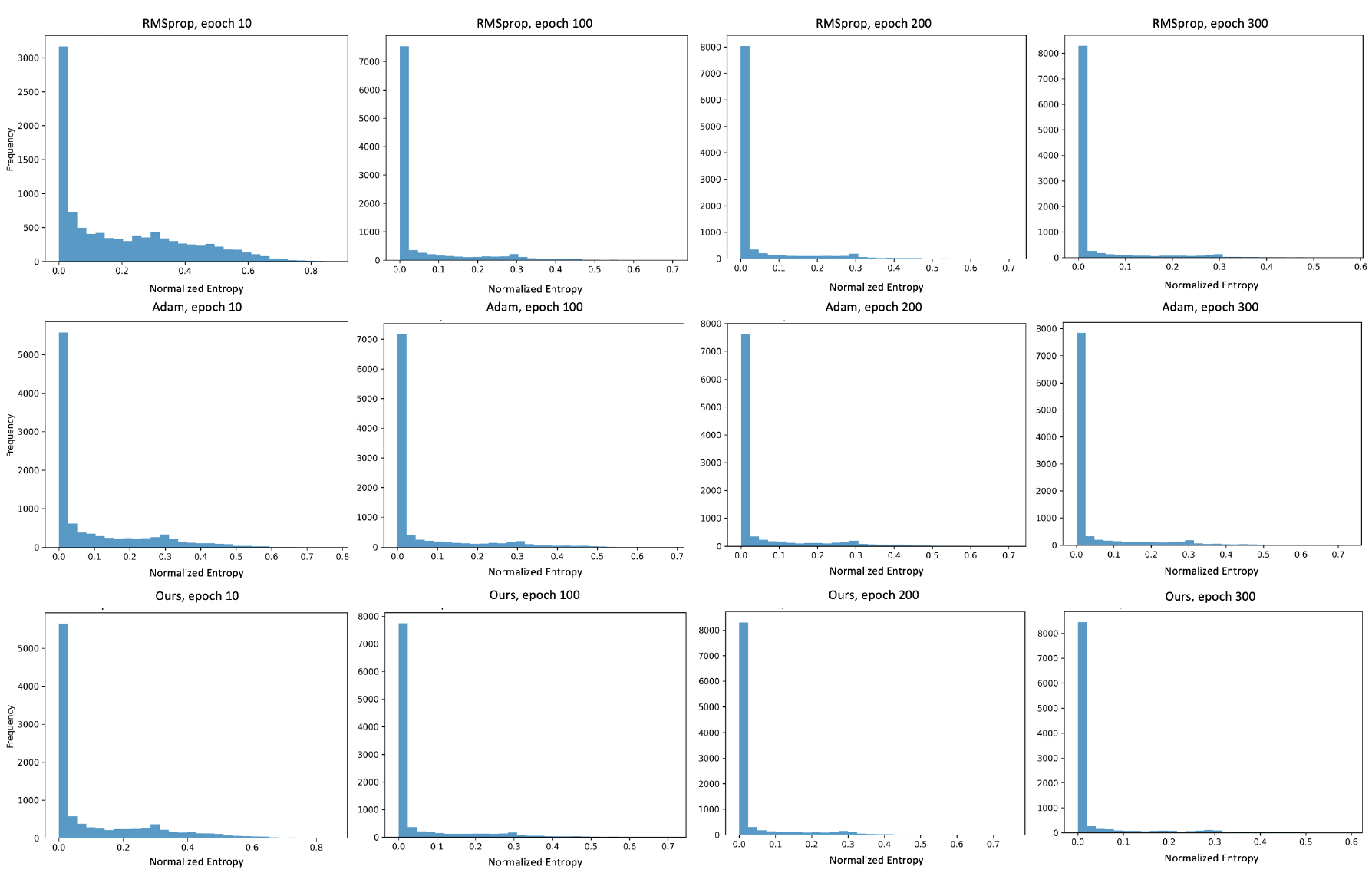}
\caption{Comparison of the distribution of normalized entropy across different optimizers (RMSProp, Adam, and StochGradAdam) at various training epochs (10, 100, 200, and 300). The histograms depict the frequency of a specific range of normalized entropy values, illustrating how the uncertainty in predictions evolves as training progresses.}
\label{fig:RMS}
\end{figure*}

Figure \ref{fig:RMS} showcases contrasting characteristics between RMSProp, Adam, and StochGradAdam (our proposed optimizer) in their approach to prediction uncertainty throughout the epochs. Initially, each optimizer displays a wide distribution in normalized entropy, reflecting a mixture of prediction confidences. However, by the 100th epoch, a distinguishing feature of StochGradAdam becomes apparent. Its histogram is markedly skewed towards the lower normalized entropy values, implying that a significant portion of its predictions are made with high confidence. RMSProp and Adam, on the other hand, also depict enhancements, but they do not attain the same degree of prediction certainty as rapidly as StochGradAdam.

\begin{figure*}[hbt!]
\centering
\includegraphics[width=1\columnwidth]{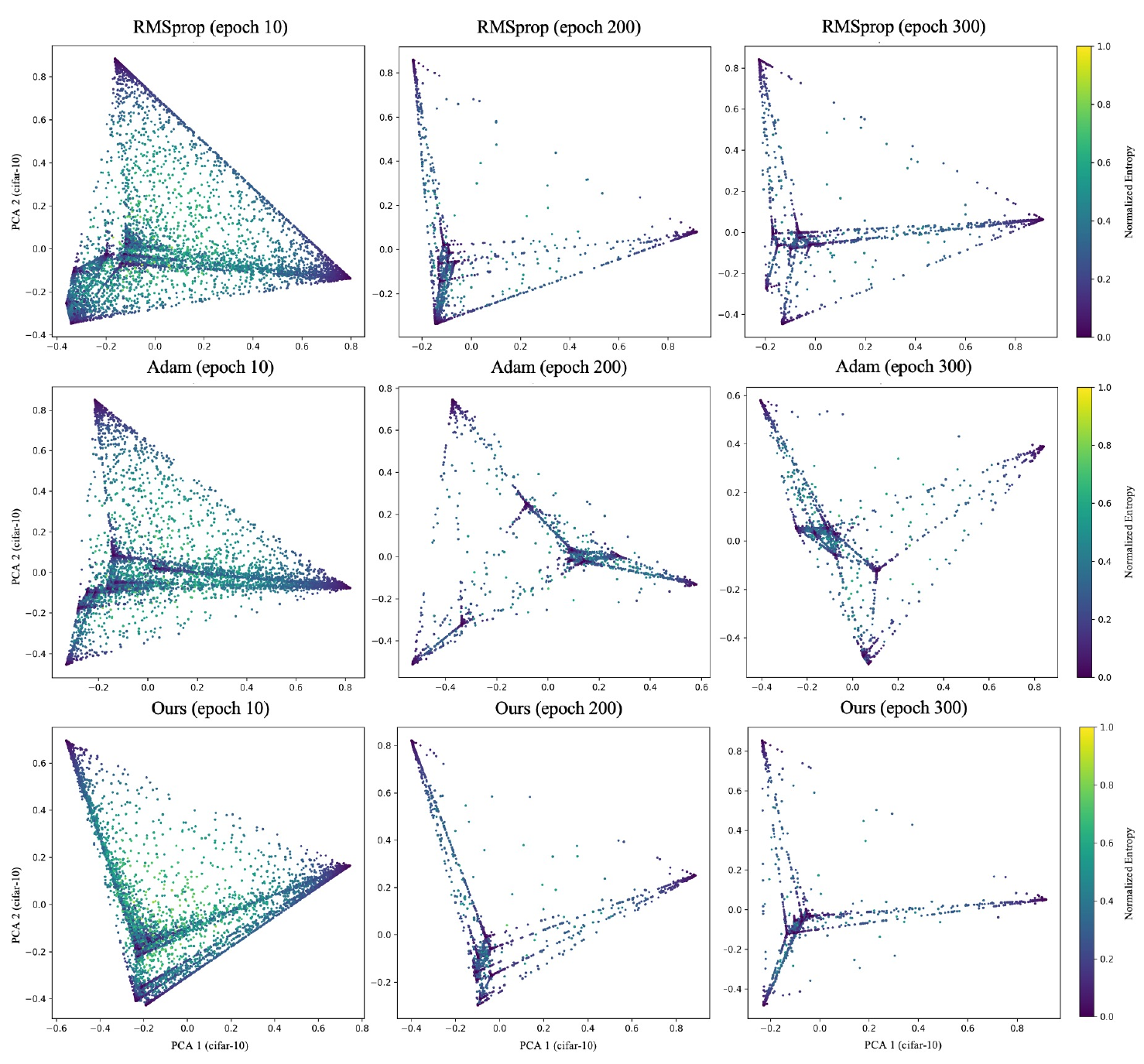}
\caption{PCA visualization of data processed with different optimizers at distinct training epochs. Each plot captures the distribution of data points in the reduced dimensional space, with color gradients representing normalized entropy.}
\label{fig:mat}
\end{figure*}

As training reaches 200 and 300 epochs, StochGradAdam consistently maintains its superior performance, ensuring its predictions are more confident. On the other hand, both RMSProp~\cite{tieleman2012rmsprop} and Adam~\cite{kingma2014adam} also improve, but their histograms still show a wider spread of entropy values, suggesting some remaining uncertainties in their predictions. StochGradAdam is particularly effective at quickly reducing prediction uncertainty, allowing models trained with it to make more confident predictions faster than those trained with RMSProp or Adam. This makes StochGradAdam potentially a better choice in situations where fast and reliable predictions are needed, such as real-time systems or applications with strict time constraints.

\subsection{Comparative Visualization: PCA}
Visual representations are crucial for understanding the differences between optimizers. These visualizations are especially helpful when evaluating their performance on benchmark datasets and architectures. For example, the performance of ResNet56~\cite{he2016deep} on CIFAR-10~\cite{krizhevsky2009learning}, trained over 300 epochs with a learning rate of 0.01, can be explored using Principal Component Analysis (PCA) to reduce the dimensions and highlight important patterns~\cite{jolliffe2016principal}. Figure \ref{fig:mat} shows a side-by-side comparison of the uncertainty landscape in neural networks, evaluated on 10,000 test data points after training.

At the beginning, around the 10th epoch, all optimizers—RMSProp~\cite{tieleman2012rmsprop}, Adam~\cite{kingma2014adam}, and Ours—show scattered data points across the triangle, indicating that they are still exploring the feature space and trying to find the best direction for optimization. RMSProp at this stage has a denser cluster near the vertex, with some points spread out, suggesting that some parts of the data have started following an optimization path, while most are still searching. Adam's points are slightly shifted to the left, indicating that it may be converging to specific areas in the feature space early on. In contrast, our method already shows a concentration of points near the base of the triangle, implying that it has begun identifying the optimal path more effectively than the other two optimizers.

By the 200th epoch, the differences between the optimizers become more noticeable. RMSProp still shows a wider distribution of points, which indicates progress but also some distance from optimal convergence. Adam's points are more tightly grouped around the center, reflecting its ability to reduce prediction uncertainty, though not evenly across all data points. In contrast, our method shows a tight cluster near the bottom of the triangle, suggesting higher prediction confidence. This indicates that our optimizer not only reduces prediction uncertainty more effectively but also achieves more consistent results, with stronger clustering.

By the 300th epoch, our method shows a compact cluster of points, indicating minimal prediction uncertainty and reinforcing the idea that dense clustering often translates to successful real-world performance. RMSProp and Adam show improvement but do not achieve the same level of clustering or confidence as our method, highlighting the superior ability of StochGradAdam to efficiently navigate the optimization landscape. The visualizations across epochs not only show the progress of each optimizer but also highlight our method's clear advantage, especially in the later stages, where it reduces prediction uncertainty more quickly and suggests faster convergence.


\section{Experimental Results}
In this section, we examine the practical performance of our proposed methodologies through a series of experiments. The primary goal is to validate our theoretical claims and assess the efficiency, robustness, and scalability of the techniques in real-world applications. Each experiment is designed to address specific performance questions and compare our methods against existing benchmarks, providing a thorough evaluation of their strengths and potential limitations.

\subsection{Image Classification: CIFAR-10}
For the image classification task, we utilized the CIFAR-10 dataset~\cite{krizhevsky2009learning}, a well-known benchmark in machine learning. We tested several neural network architectures, including ResNet~\cite{he2016deep}, VGG16~\cite{simonyan2014very}, MobileNetV2~\cite{mobilenetv2}, and Vision Transformer~\cite{dosovitskiy2020image}, to compare their performance using different optimizers, RMSProp~\cite{tieleman2012rmsprop} and Adam~\cite{kingma2014adam}. Each model was trained at a learning rate of 0.01, sampling rate 0.8 (using 80\% from entire gradients) for StochGradAdam, 128 of batch size, and 300 epochs using TensorFlow.

\begin{figure*}[hbt!]
\centering
\includegraphics[width=1\columnwidth]{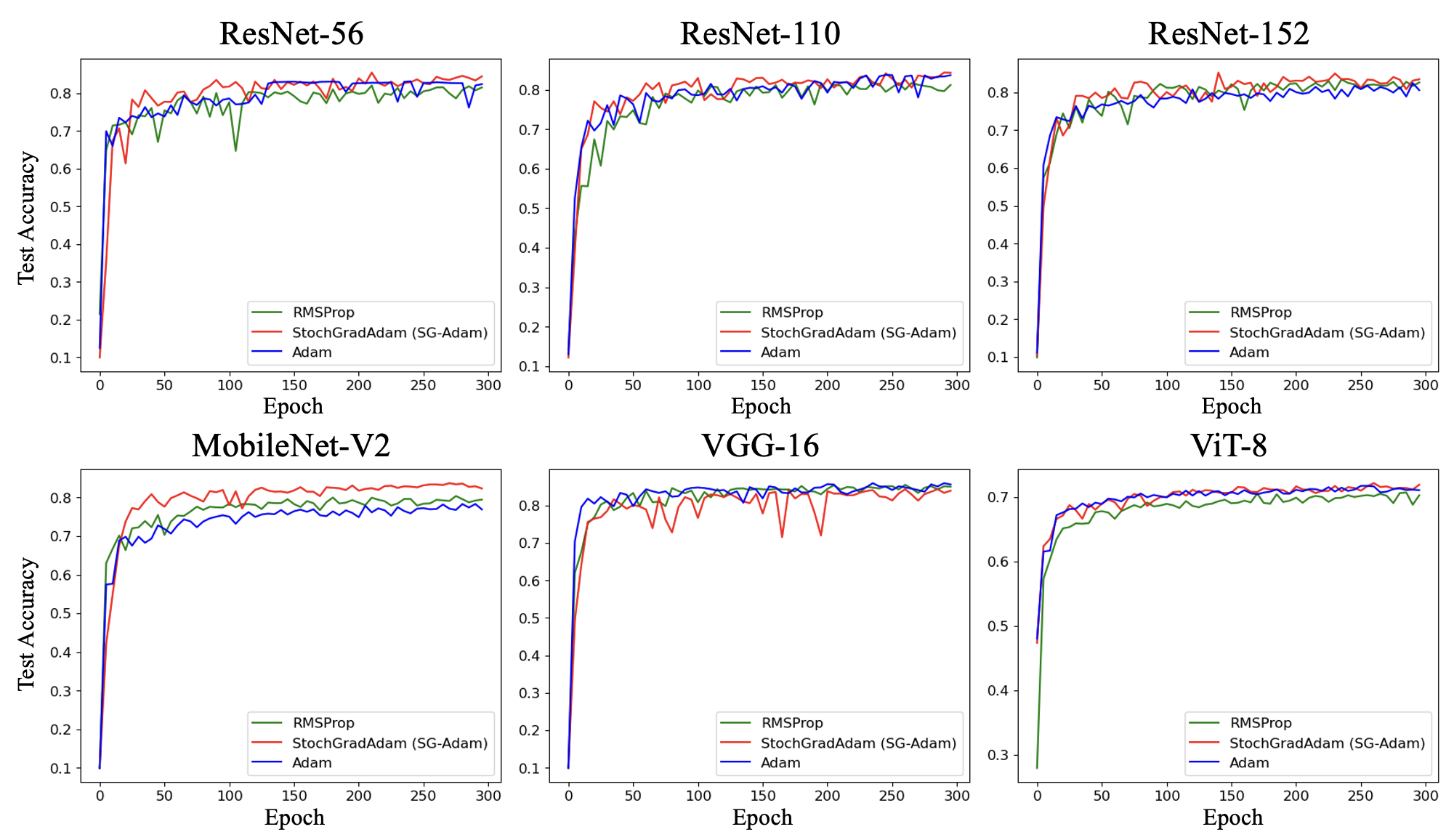}
\caption{Comparison of test accuracy over 300 epochs on CIFAR-10 dataset~\cite{krizhevsky2009learning} for various neural network architectures: ResNet-56,110,152~\cite{he2016deep}, MobileNetV2, VIT-8~\cite{dosovitskiy2020image}, and VGG-16~\cite{simonyan2014very}. Three different optimizers - RMSprop (green)~\cite{tieleman2012rmsprop}, Adam (blue)~\cite{kingma2014adam}, and StochGradAdam (red) - were used to train each model.}
\label{fig:tacc}
\end{figure*}

\setlength{\tabcolsep}{8pt} 
\begin{table}[htbp]
\centering
\footnotesize
\caption{Top-1 Accuracy and Training Loss on CIFAR-10~\cite{krizhevsky2009learning} over 300 epochs for various neural network architectures: ResNet-56, 110, 152~\cite{he2016deep}, MobileNetV2~\cite{mobilenetv2}, VIT-8~\cite{dosovitskiy2020image}, and VGG-16~\cite{simonyan2014very}.}
\begin{adjustbox}{width=1\textwidth}
\begin{tabular}{@{}lcccccc@{}}
\toprule
& \multicolumn{6}{c}{Performance: Top 1 Accuracy} \\
\cmidrule(lr){2-3}
\cmidrule(lr){4-5}
\cmidrule(lr){6-7}
 & \multicolumn{2}{c}{RMSProp~\cite{tieleman2012rmsprop}} & \multicolumn{2}{c}{Adam~\cite{kingma2014adam}}& \multicolumn{2}{c}{StochGradAdam (Ours)} \\
\cmidrule(lr){2-7} 
Architectures & Test Acc. & Train Loss. & Test Acc. & Train Loss. & Test Acc. & Train Loss. \\
\midrule
ResNet-56~\cite{he2016deep} & 0.824 & 2.2e-03 & 0.831 & 9.9e-05 & \textbf{0.853} & 1.2e-03  \\
ResNet-110~\cite{he2016deep} & 0.824 & 3.2e-03 & 0.838 & 8.4e-05 & \textbf{0.847} & 3.9e-03  \\
ResNet-152~\cite{he2016deep} & 0.833 & 3.8e-03 & 0.827 & 1.9e-04 & \textbf{0.851} & 1.0e-03  \\
MobileNetV2~\cite{mobilenetv2} & 0.803 & 3.6e-03 & 0.783 & 7.0e-03 & \textbf{0.840} & 3.3e-03  \\
VGG-16~\cite{simonyan2014very} & 0.855 & 2.2e-03 & \textbf{0.861} & 3.7e-04 & 0.843 & 3.9e-03  \\
Vision Transformer-8~\cite{dosovitskiy2020image} & 0.709 & 5.4e-02 & 0.718 & 6.7e-02 & \textbf{0.721} & 7.7e-02  \\
\bottomrule
\end{tabular}

\label{tab:bfloat}
\end{adjustbox}
\end{table}

Figure~\ref{fig:tacc} presents a graphical comparison of test accuracy for different deep learning architectures when trained with three optimizers: RMSprop~\cite{tieleman2012rmsprop}, Adam~\cite{kingma2014adam}, and our StochGradAdam.

The ResNet-56 model demonstrates that StochGradAdam outperforms both RMSProp and Adam, achieving a higher test accuracy. The graph shows that StochGradAdam quickly rises in accuracy during the early epochs and maintains a consistent lead over the other optimizers, reaching a final test accuracy of 0.853.

Similarly, for the ResNet-110 model, StochGradAdam shows a strong initial performance and sustains a higher accuracy throughout the training process. Its final test accuracy of 0.847 is superior to both RMSProp and Adam.

In the ResNet-152 model, StochGradAdam exhibits a rapid rise in accuracy during the early stages of training, achieving a final accuracy of 0.851. Although the performance of all optimizers converges towards the end, StochGradAdam consistently shows better results.

For the MobileNetV2 model, StochGradAdam demonstrates superior performance compared to the other optimizers. The graph indicates that StochGradAdam achieves a higher test accuracy across the entire training duration, reaching 0.840.

In the case of the VGG-16 model, StochGradAdam shows more variability in its performance compared to Adam. The final test accuracy of StochGradAdam is 0.843, slightly lower than Adam’s 0.861. This result suggests that for the VGG-16 architecture, StochGradAdam may not perform as well as it does for other architectures.

For the Vision Transformer (ViT-8), StochGradAdam maintains stable progress, performing better than RMSProp and reaching a final test accuracy of 0.721. Although its performance is similar to Adam’s, StochGradAdam still holds a slight advantage in the final results.

In summary, StochGradAdam consistently demonstrates better performance across most architectures, particularly excelling in the early stages of training. However, its performance in the VGG-16 architecture shows some limitations, suggesting that further investigation may be needed to enhance its adaptability to certain model types.

\subsection{Image Segmentation} \JY{I need to fix and add more datasets for image segmentation}
Building on our previous discussion of classification, we moved into another key area of deep learning: segmentation. For this experiment, we used the Unet-2 architecture \cite{ronneberger2015u} combined with MobileNetV2 \cite{mobilenetv2}, striking a balance between computational efficiency and performance. We used StochGradAdam as the optimizer for these experiments, with a learning rate of 0.001.

\begin{figure*}[hbt!]
\centering
\includegraphics[width=1\columnwidth]{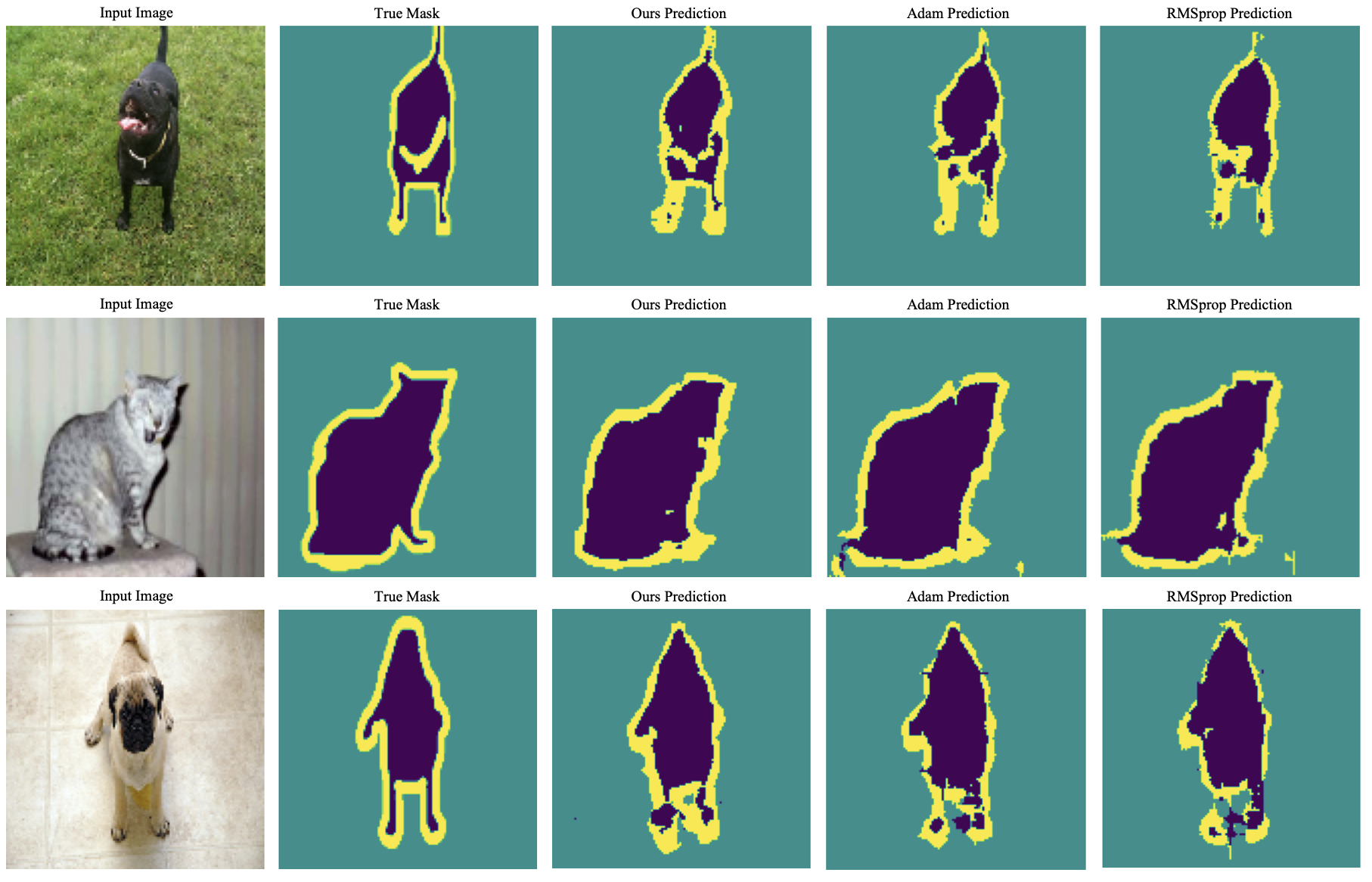}
\caption{A comparative visualization of segmentation results on the oxford\_iiit\_pet dataset using the Unet-2~\cite{ronneberger2015u} architecture combined with MobileNetV2~\cite{mobilenetv2}. The figure compares results across different optimizers, including StochGradAdam, Adam, and RMSProp.}
\label{fig:tacc2}
\end{figure*}

The dataset we used is the well-known oxford\_iiit\_pet \cite{oxford_iiit_pet}, which contains a strong collection of images suitable for segmentation tasks. As seen in Fig. \ref{fig:tacc2}, the results of our experiments highlight the effectiveness of StochGradAdam. The column labeled "Ours Prediction" shows more accurate and consistent segmentation compared to other commonly used optimizers such as Adam and RMSProp. The edges of the segments are clearer, and the segmentation masks more closely match the actual object boundaries, showing the ability of StochGradAdam to improve model performance.

In conclusion, StochGradAdam proved to be not only effective for classification tasks but also highly capable in segmentation tasks. The results, supported by the visual comparisons, suggest that StochGradAdam sets a higher standard for future optimizers in this field.

\section{Discussion}
In our exploration of our optimizer, we've delved deep into its intricacies, nuances, and potential advantages in the realm of neural architectures. The results garnered from various architectures illuminate not just the merits of our approach but also the subtleties of how different neural architectures respond to gradient manipulations. As with any methodological advance, while the advantages are manifold, it is crucial to be cognizant of the boundaries and constraints. Before presenting our conclusive thoughts on the methodology, it is pertinent to discuss the limitations observed during our study. The understanding of these constraints not only provides clarity about the method's scope but also lays the groundwork for potential future improvements.

\subsection{Hyperparameter tuning: Sample Rate}
n our experiments, we varied the sample rate to analyze its impact on model performance across different ResNet architectures (ResNet-56, ResNet-110, and ResNet-152). The sample rate, in this context, refers to the percentage of gradients used during the update step of the StochGradAdam optimizer. Specifically, we tested with sample rates of 100\%, 80\%, 60\%, and 20\%.

\begin{figure*}[hbt!] 
\centering 
\includegraphics[width=1\columnwidth]{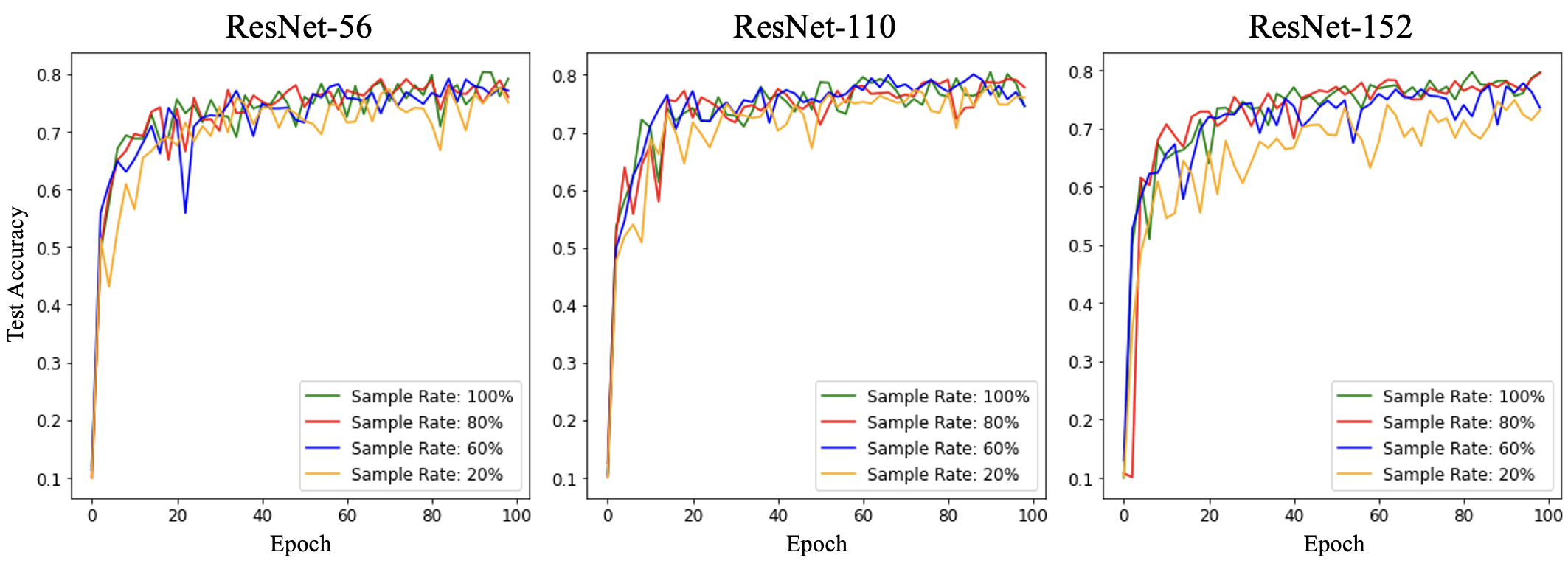} 
\caption{Test accuracy across different sample rates (100\%, 80\%, 60\%, and 20\%) for ResNet-56, ResNet-110, and ResNet-152 architectures~\cite{he2016deep}. The sample rate refers to the percentage of gradients used during updates. All experiments are done with Adam and StochGradAdam on CIFAR-10~\cite{krizhevsky2009learning} with 256 of batch size and 0.001 of learning rate} 
\label{fig:tune}
\end{figure*}

As depicted in Figure \ref{fig:tune}, we observed that the models trained with higher sample rates, such as 100\% and 80\%, consistently achieved better and more stable performance compared to those trained with lower sample rates, such as 60\% and 20\%. This is particularly evident in the early stages of training, where the models with lower sample rates demonstrated slower convergence and higher variability in test accuracy. The models trained with a 100\% sample rate (i.e., using all gradients - just normal Adam~\cite{kingma2014adam}) displayed rapid convergence and reached a higher final test accuracy in fewer epochs.

Interestingly, despite the significant drop in sample rate to just 20\%, the models still maintained some accuracy. While there was a noticeable dip in test accuracy across ResNet-56, ResNet-110, and ResNet-152 when only 20\% of the gradients were used for updates, the models were still able to reach fairly reasonable accuracy levels. This suggests that even with a substantial reduction in gradient information, the optimization process can still proceed effectively, though at a slower pace and with slightly more variability. This highlights the resilience of gradient-based optimization techniques, even when only a fraction of the gradients are being utilized.

In conclusion, although higher sample rates provide faster convergence and more stable performance during the initial epochs, lower sample rates, such as 20\%, can still maintain reasonable accuracy given enough training time. This highlights an intriguing property of stochastic gradient sampling, where the model can still progress effectively with reduced gradient information, albeit with slower convergence and slightly more noise.

\subsection{Limitations}
Our gradient sampling technique has shown significant success in architectures like ResNet~\cite{he2016deep} and MobileNetV2~\cite{mobilenetv2}, which incorporate mechanisms such as residual connections and bottleneck structures that effectively mitigate the vanishing gradient problem. These mechanisms help preserve the gradient flow during backpropagation. However, deeper architectures like VGG~\cite{simonyan2014very}, which lack these built-in mechanisms, pose challenges. This limitation is primarily due to the well-known vanishing gradient problem in deep architectures that do not employ gradient-preserving techniques like skip connections.

\subsubsection{Deep Gradient Vanishing: Mathematical Formulation}
In deep neural networks, the gradient \( \nabla L(\theta)^{(l)} \) at layer \( l \) is recursively computed based on the gradient at deeper layers. Specifically, the gradient at layer \( l \) is given by the following recursive relation:
\begin{equation}
\nabla L(\theta)^{(l)} = \left( \theta^{(l)} \right)^T \nabla L(\theta)^{(l+1)} \circ f'^{(l)}(z^{(l)})
\end{equation}
Where:
\begin{itemize}
    \item \( \nabla L(\theta)^{(l)} \) is the gradient of the loss function \( L(\theta) \) with respect to the weights \( \theta^{(l)} \) at layer \( l \),
    \item \( \theta^{(l)} \) is the weight matrix at layer \( l \),
    \item \( f'^{(l)}(z^{(l)}) \) represents the derivative of the activation function at the output \( z^{(l)} \) of layer \( l \),
    \item \( \circ \) denotes element-wise multiplication.
\end{itemize}
As the depth of the network increases, the product of weight matrices \( \theta^{(l)} \) and the derivatives \( f'^{(l)}(z^{(l)}) \) becomes increasingly susceptible to diminishing values. Particularly, when \( |f'^{(l)}(z^{(l)})| < 1 \) for many layers, the gradients \( \nabla L(\theta)^{(l)} \) in earlier layers decay exponentially. This leads to vanishing gradients, making it difficult for earlier layers to learn effectively.

\subsubsection{Quantitative Analysis of Gradient Decay}
Assume that for each layer \( l \), the derivative of the activation function is bounded as \( |f'^{(l)}(z^{(l)})| \leq \beta \) where \( 0 < \beta < 1 \). Then, for a network with \( L \) layers, the gradient at the first layer \( \nabla L(\theta)^{(1)} \) can be bounded as follows~\cite{he2016deep}:
\begin{equation}
|\nabla L(\theta)^{(1)}| \leq \beta^L |\nabla L(\theta)^{(L)}|
\end{equation}
As the number of layers \( L \) increases, the term \( \beta^L \) becomes exponentially small, leading to vanishing gradients in the earlier layers compared to the final layer. For large \( L \), this phenomenon severely hampers the effectiveness of training in deeper networks like VGG~\cite{simonyan2014very}.

\subsubsection{Consequences for Gradient Sampling}
Our gradient sampling technique relies on capturing and updating selective gradient components. However, in the presence of vanishing gradients, the magnitude of the gradients in earlier layers diminishes, reducing the informativeness of the gradient signals. When stochastic sampling is performed from a distribution in which most gradients have negligible magnitude, the variance of the sampled gradients increases. This is expressed as:
\begin{equation}
\text{Variance of sampled gradients} \propto \frac{1}{|\nabla L(\theta)^{(l)}|}
\end{equation}
As the gradient magnitudes decrease, the variance in the sampled gradients grows larger, leading to noisier updates and less effective optimization. This is particularly problematic in architectures that lack gradient-preserving features, such as VGG. The combination of vanishing gradients and increased variance in gradient sampling impairs the directionality and effectiveness of the updates, resulting in slower convergence or even failure to converge.

\subsubsection{Practical Implications and Future Directions}
While our gradient sampling technique has demonstrated strong performance in architectures like ResNet that mitigate the vanishing gradient problem, it faces challenges in deeper architectures that do not preserve gradient flow. In models like VGG, which lack residual connections or other gradient-preserving techniques, further refinements to the sampling method are needed. Future research could focus on adjusting the sampling process or incorporating alternative mechanisms, such as skip connections or regularization techniques, to mitigate these issues in deep networks.


\section{Conclusion}
In the realm of deep learning optimization, the introduction of the StochGradAdam optimizer marks a significant stride forward. Central to its design is the innovative gradient sampling technique, which not only ensures stable convergence but also potentially mitigates the effects of noisy or outlier data. This approach fosters robust training and enhances the exploration of the loss landscape, leading to more dependable convergence.

Throughout the empirical evaluations, StochGradAdam consistently demonstrated superior performance in various tasks, from image classification to segmentation. Especially noteworthy is its ability to reduce prediction uncertainty, a facet that goes beyond mere accuracy metrics. This reduction in uncertainty is indicative of the model's robustness and its potential resilience against adversarial attacks and noisy data.

However, like all methodologies, StochGradAdam has its limitations. While it excels in architectures like ResNet and MobileNet, challenges arise in deeper architectures like VGG, which lack certain mitigating features. This limitation is believed to be rooted in the vanishing gradient problem, prevalent in deep architectures without protective mechanisms.

Nevertheless, the successes of StochGradAdam underscore the potential for further innovation in gradient-based optimizations. Its rapid convergence, adaptability across diverse architectures, and ability to reduce prediction uncertainty set a new benchmark for future optimizers in deep learning.

\bibliographystyle{plain}
\bibliography{main}

\end{document}